\documentclass{article}
\usepackage{balance}
\usepackage{flushend}
\usepackage{spconf,amsmath,amssymb,graphicx,hyperref}
\usepackage{booktabs}   
\usepackage{colortbl}   
\usepackage{xcolor}     
\usepackage{multirow}
\usepackage{etoolbox}
\makeatletter
\apptocmd{\thebibliography}{%
  \setlength{\itemsep}{0pt}
  \setlength{\parskip}{0pt}
  \setlength{\parsep}{0pt}%
}{}{}
\makeatother

\title{Boot-and-Feedback Framework for Generalist-Expert Model Collaboration in Breast Ultrasound Diagnosis}

\name{\parbox{\linewidth}{\centering
Ming Cheng$^{1}$, Hongyu Sun$^{2}$, Zhaolin Chen$^{1}$, Jun Liu$^{3}$, Hossein Rahmani$^{3}$, Qiuhong Ke$^{1\dagger}$
}
\thanks{
  \begin{tabular}{@{}l@{}}
    $^\dagger$Corresponding author
  \end{tabular}
    }
}
\address{$^1$ Department of Data Science \& AI, Monash University, Australia \\
$^2$ Department of Computer Science, Renmin University of China, China \\
$^3$ School of Computing and Communications, Lancaster University, UK }

\usepackage{caption}

\renewcommand{\thetable}{\Roman{table}}
\begin{document}
%
\maketitle

%
\begin{abstract}
Breast ultrasound (BUS) is widely used for breast cancer diagnosis yet remains operator-dependent. While deep learning shows promise, ensuring diagnostic reliability and interpretability is challenging. Recent Multimodal Large Language Models (MLLMs) often generate spurious descriptions due to limited domain knowledge, which mislead downstream expert models and compromise clinical validity. To address these challenges, we propose the \textbf{Boo}t-and-\textbf{F}eedback (BooF) model collaboration framework for synergistic MLLM-expert interaction. Specifically, in the Boot Stage, the MLLM is guided by the BI-RADS lexicon and preliminary benign-malignant vision-expert predictions, enabling it to transfer general reasoning to BUS analysis while avoiding hallucinations. Subsequently, the Feedback Stage integrates these descriptions with visual features via a lightweight Attention-Gated Cross-Modality Fusion Module. This allows the expert to leverage textual feedback while adaptively filtering noise. Extensive experiments on multiple BUS datasets demonstrate that BooF substantially outperforms state-of-the-art methods in terms of diagnostic accuracy and interpretability.
\end{abstract}
%

\section{Introduction}
\label{sec:intro}
\begin{figure}[!t]
    \centering
    \setlength{\belowcaptionskip}{0pt} 
    \includegraphics[width=1\columnwidth]{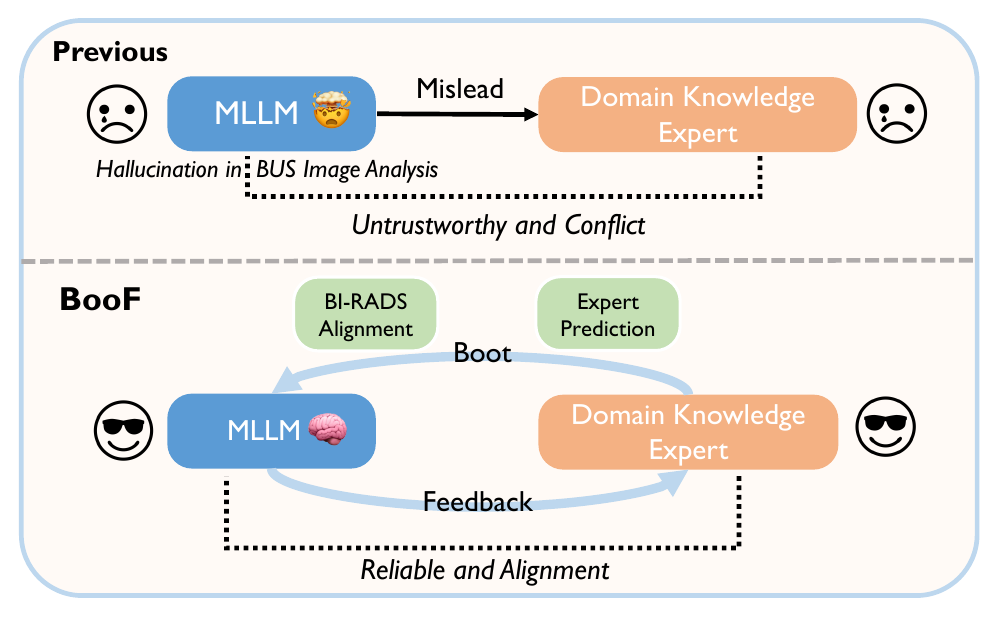}
\vspace{-15pt}
    \caption{Comparison between Boot-and-Feedback Framework and previous unidirectional model interaction.}
\vspace{-20pt}
\label{fig:small1}
\end{figure}

\begin{figure*}[t]
\centering
\includegraphics[width=\textwidth]{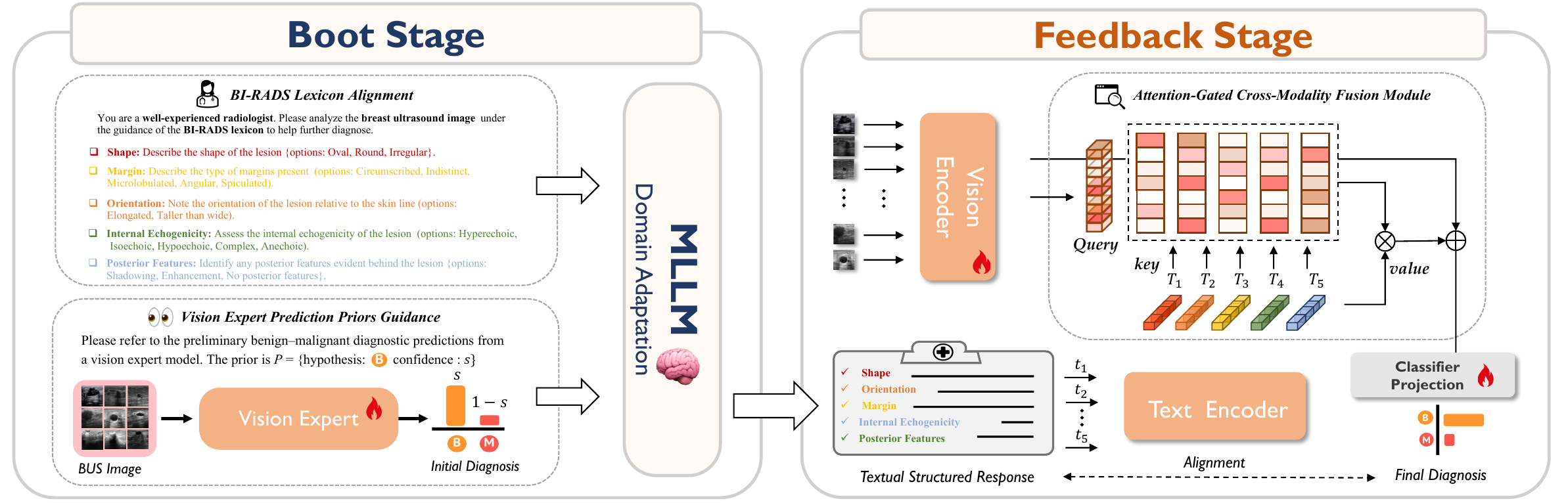}
\vspace{-20pt}
\caption{Overview of the proposed Boot-and-Feedback (BooF) framework. 
The Boot Stage adapts the MLLM with BI-RADS lexicon alignment and vision-expert priors to generate structured lesion descriptions. 
The Feedback Stage fuses these descriptions with visual features via an attention-gated module for the final diagnosis.}
\vspace{-15pt}
\label{main_figure}
\end{figure*}
 
Breast cancer is the most prevalent malignancy among women, and early detection is crucial \cite{WHO}. Breast ultrasound (BUS) is a widely used diagnostic technique due to its non-invasiveness and accessibility, but its interpretation remains operator-dependent, causing variability across clinical settings \cite{kelly2010}. This challenge has driven the development of computer-aided diagnosis (CAD) systems that provide accurate, interpretable results to assist radiologists.
\newline\indent
Although deep learning has advanced, it has not yet produced an ideal CAD system that ensures both high accuracy and interpretability. Domain-specific models based on CNNs \cite{resnet} and ViTs \cite{vit} show competitive performance in breast cancer diagnosis but lack structured clinical reasoning, limiting interpretability. Multimodal large language models (MLLMs) offer a promising alternative by generating human-readable diagnostic descriptions \cite{zhang2023biomedclip, MedAlign2023}, but their application in BUS is limited by hallucinations and domain misalignment due to the lack of medical priors.

Recent studies \cite{gao2024aligning, fang2024aligning, dai2026cinedub} have leveraged MLLM-generated descriptions to enhance expert model performance, but our preliminary experiments in Table \ref{tab:ablation} indicate that such approaches often fail or degrade performance. Critically, MLLM-generated outputs frequently conflict with expert model predictions, undermining reliability in clinical applications. As shown in Figure \ref{fig:small1}, these issues arise from two main factors: (i) the lack of domain-specific knowledge in MLLMs, which leads to hallucinations; and (ii) the unidirectional interaction between MLLMs and expert models, which limits feedback and accuracy.

To address these challenges, we propose a \textbf{Boo}t-and-\textbf{F}eedback (BooF) model collaborative framework, which enables bidirectional interaction between general MLLM and domain-specific expert models. In the boot stage, we propose two MLLM boot strategies: (1) BI-RADS Lexicon Alignment. This strategy bridges the MLLM’s general image analysis capability with standardized diagnostic terminology and constrains the model to produce BI-RADS–based descriptions instead of direct diagnostic judgments. (2) Vision Expert Prediction Priors Guidance. Compared with unidirectional flow, this strategy incorporates preliminary benign–malignant predictions from a Vision Expert to serve as a reference for the MLLM, thereby suppressing hallucinations and ensuring clinically consistent diagnostic outputs. In the feedback stage, we feed the enhanced MLLM-generated descriptions together with the ultrasound images into a Vision–Language Expert to further improve classification performance. Specifically, we design a lightweight Attention-Gated Cross-Modality Fusion Module that adaptively filters potential semantic hallucination noise to ensure reliable diagnostic predictions.\newline\indent
Through extensive ablation studies, we demonstrate the effectiveness of each proposed component. Comparisons on two public datasets show that BooF outperforms state-of-the-art methods in both accuracy and interpretability.
\section{METHODS}
\subsection{Boot Stage: MLLM Domain Adaptation}
\subsubsection{BI-RADS Lexicon Alignment}
\vspace{-5pt}
In early experiments, we directly applied Gemini 2.5 Pro \cite{comanici2025gemini}, a state-of-the-art MLLM, to the breast ultrasound (BUS) diagnosis task. However, without domain adaptation, the model failed to generalize from generic image analysis to BUS interpretation, frequently producing hallucinatory diagnostic reasoning rather than evidence-based inference. This not only undermines clinical validity but also degrades downstream Vision–Language Expert performance compared with single-modality baselines (Table~\ref{tab:ablation}). This failure stems from an inherent property of MLLMs: although they are proficient at describing general image semantics, they struggle with domain-specific reasoning and clinical decision making.\newline\indent
To overcome these limitations, we reformulated the task by mirroring radiologists’ reasoning, which integrates BI-RADS criteria with imaging characteristics. The Breast Imaging Reporting and Data System (BI-RADS) \cite{BIRADS} provides a standardized lexicon for describing five key lesion properties—\emph{shape}, \emph{orientation}, \emph{margin}, \emph{internal echogenicity}, and \emph{posterior features}. Instead of soliciting direct benign/malignant predictions, we therefore guided the MLLM to generate BI-RADS–aligned lesion descriptions.
This approach ensures clinical validity and consistency while delegating the final classification to a downstream vision language expert. Specifically, we employ \emph{task-activating prompting} \cite{yang2023generative} to role-play the MLLM as an experienced radiologist, thereby activating medical reasoning, and use \emph{in-context learning (ICL)} \cite{brown2020language} to constrain outputs to the BI-RADS attribute space.\newline\indent
In addition, we establish explicit correspondences between the BI-RADS descriptors and the low-level image features, for example, linking \emph{orientation} with the spatial distribution of the lesion and \emph{internal echogenicity / posterior features} with the grayscale intensity patterns. This bridging mechanism enables the MLLM to better transfer its general visual competence into the BUS domain, producing descriptions that are semantically valid and grounded in visual evidence.
\vspace{-5pt}
\subsubsection{Vision Expert Prediction Priors Guidance}
\vspace{-5pt}
Although the BI-RADS Alignment Strategy introduces domain priors and partially mitigates hallucinations, its performance remains insufficient for clinical reliability. More importantly, the lack of bidirectional interaction leads MLLMs to produce reasoning misaligned with downstream multimodal predictions, resulting in inconsistent diagnostic outcomes.

Although fine-tuning could address these issues, it is prohibitively expensive and infeasible for closed-source MLLMs. To this end, we propose the Vision Expert Prediction Priors (VEPP) Strategy, which leverages a lightweight vision expert trained on BUS images. During inference, the expert provides probabilistic benign--malignant estimates $p \in [0,1]$. After confidence calibration, we compute $s = \max(p, 1-p)$. If $s \geq \tau$ (with $\tau$ selected on the validation set), we construct a structured prior $\mathcal{P} = \{\texttt{hypothesis}: \texttt{benign/malignant},\ \texttt{confidence}: s\}$. Otherwise, no prior is provided. The prior is injected into the boot-stage prompt alongside the BI-RADS template. The instructions emphasize: \emph{``Use $\mathcal{P}$ only as a reference; generate BI-RADS--aligned descriptors, and explicitly state contradictions if the image evidence disagrees.''}. This design reframes the open-ended query of \emph{``Is there cancer?''} into a constrained, hypothesis-driven description task.

By narrowing the reasoning space while preserving independence, VEPP reduces hallucination-prone outputs and allows the MLLM to challenge erroneous priors. Empirically, we observe that even when the expert's initial prediction is wrong, the MLLM can flag inconsistencies under BI-RADS guidance and produce coherent descriptions, enabling the downstream vision-language expert to recover the correct classification. Overall, VEPP improves robustness, aligns textual reasoning with classifier decision space, and enhances clinical reliability.
\vspace{-5pt}
\subsection{Feedback Stage: Multimodal Expert Enhancement}
\subsubsection{Vision-Language Expert Model Pipeline}
\vspace{-5pt}
In the feedback stage, MLLM-generated BI-RADS lesion descriptions are integrated into a downstream Vision–Language Expert, establishing a bidirectional interaction with the boot stage. Unlike the boot stage, which relies solely on vision cues, this vision-language expert model jointly encodes textual descriptors and BUS image features to refine the prediction of malignancy.

Concretely, each BI-RADS attribute description $t_i$ is encoded as $\mathbf{T}_i = f_T(t_i)$ using a frozen text encoder, where $i \in {1,2,\dots,5}$ corresponds to the five BI-RADS properties. The text encoder is kept frozen to ensure the preservation of standardized semantics. The original BUS image $X$ is processed with a trainable image encoder $f_I(\cdot)$, producing a global feature representation $\mathbf{I} = f_I(X)$. Both encoders are initialized from large-scale pre-trained models, enabling efficient transfer with limited annotated BUS images \cite{dai2023improving}. \newline\indent
As a result, this bidirectional framework reinforces the concordance between diagnostic predictions and generated descriptions, thereby improving clinical interpretability.  By analogy to the boot stage, an optional extension is to propagate the predictions of the vision language expert back into the MLLM, which can further strengthen the alignment between classification outcomes and textual descriptions.

\begin{table*}[tbp]
\centering
\scriptsize
\caption{Ablation on BUS-BRA \cite{gomez2024bus} with two backbones. Values are Mean $\pm$ Std over 5 folds. Best results per backbone are \textbf{bold}.}
\vspace{-8pt}
\resizebox{\textwidth}{!}{
\begin{tabular}{llcccccc}
\toprule[1pt]
\textbf{Backbone} & \textbf{Method} & \textbf{AUC$\uparrow$} & \textbf{Accuracy$\uparrow$} & \textbf{Specificity$\uparrow$} & \textbf{Precision$\uparrow$} & \textbf{Recall$\uparrow$} & \textbf{F1-score$\uparrow$} \\
\midrule
\multirow{5}{*}{\textbf{MedViT \cite{medvit}}}
& Vision expert                     & 0.915 $\pm$ 0.017 & 0.860 $\pm$ 0.017 & 0.906 $\pm$ 0.024 & 0.796 $\pm$ 0.042 & 0.759 $\pm$ 0.053 & 0.776 $\pm$ 0.039 \\
& Vision-language expert                    & 0.919 $\pm$ 0.015 & 0.862 $\pm$ 0.015 & 0.898 $\pm$ 0.022 & 0.787 $\pm$ 0.035 & 0.786 $\pm$ 0.005 & 0.786 $\pm$ 0.018 \\
& \ +BI-RADS lexicon                & 0.932 $\pm$ 0.010 & 0.880 $\pm$ 0.008 & 0.933 $\pm$ 0.009 & 0.845 $\pm$ 0.028 & 0.769 $\pm$ 0.025 & 0.805 $\pm$ 0.019 \\
&  \ \ + Prior guidance  & 0.975 $\pm$ 0.006 & 0.926 $\pm$ 0.008 & 0.942 $\pm$ 0.012 & 0.881 $\pm$ 0.021 & 0.891 $\pm$ 0.029 & 0.886 $\pm$ 0.018 \\
&  \ \ \ + Attention-gated
                                   & \textbf{0.976 $\pm$ 0.004} & \textbf{0.934 $\pm$ 0.005} & \textbf{0.949 $\pm$ 0.006} & \textbf{0.895 $\pm$ 0.010} & \textbf{0.903 $\pm$ 0.019} & \textbf{0.899 $\pm$ 0.011} \\
\midrule
\multirow{5}{*}{\textbf{ResNet-50 \cite{resnet}}}
& Vision expert                      & 0.927 $\pm$ 0.012 & 0.869 $\pm$ 0.011 & 0.933 $\pm$ 0.012 & 0.839 $\pm$ 0.034 & 0.734 $\pm$ 0.033 & 0.783 $\pm$ 0.025 \\
& Vision-language expert                     & 0.919 $\pm$ 0.008 & 0.861 $\pm$ 0.006 & 0.939 $\pm$ 0.023 & 0.848 $\pm$ 0.047 & 0.696 $\pm$ 0.046 & 0.762 $\pm$ 0.016 \\
& \ +BI-RADS lexicon                 & 0.931 $\pm$ 0.013 & 0.877 $\pm$ 0.015 & 0.932 $\pm$ 0.019 & 0.843 $\pm$ 0.042 & 0.764 $\pm$ 0.049 & 0.800 $\pm$ 0.027 \\
& \ \ + Prior guidance        & 0.969 $\pm$ 0.006 & 0.916 $\pm$ 0.012 & 0.937 $\pm$ 0.018 & 0.871 $\pm$ 0.035 & 0.873 $\pm$ 0.023 & 0.871 $\pm$ 0.018 \\
& \ \ \ + Attention-gated
                                   & \textbf{0.975 $\pm$ 0.004} & \textbf{0.926 $\pm$ 0.008} & \textbf{0.943 $\pm$ 0.017} & \textbf{0.883 $\pm$ 0.030} & \textbf{0.891 $\pm$ 0.018} & \textbf{0.887 $\pm$ 0.015} \\
\bottomrule[1pt]
\end{tabular}}
\label{tab:ablation}
\vspace{-8pt}
\end{table*}

\begin{table}[tbp]
\centering
\scriptsize
\caption{Quantitative comparison on BUSI \cite{BUSI} and BUS-BRA \cite{gomez2024bus}. Values are Mean $\pm$ Std over 5 folds. Best results are \textbf{bold}.}
\vspace{-8pt}
\setlength{\tabcolsep}{3pt} 
\renewcommand{\arraystretch}{1.05} 
\begin{tabular}{lcccc}
\toprule[1pt]
\multicolumn{5}{c}{\textbf{BUSI}} \\
\midrule
\textbf{Method} & \textbf{AUC$\uparrow$} & \textbf{Specificity$\uparrow$} & \textbf{Recall$\uparrow$} & \textbf{F1-score$\uparrow$} \\
\midrule
KRC-APM \cite{krc}       & 0.955 $\pm$ 0.005 & 0.894 $\pm$ 0.009 & 0.847 $\pm$ 0.033 & 0.864 $\pm$ 0.021 \\
HoverTrans \cite{hovertrans}   & 0.957 $\pm$ 0.010 & 0.918 $\pm$ 0.016 & 0.858 $\pm$ 0.028 & 0.864 $\pm$ 0.028 \\
REAF \cite{reaf}         & 0.955 $\pm$ 0.004 & 0.929 $\pm$ 0.018 & \textbf{0.868} $\pm$ 0.035 & 0.880 $\pm$ 0.013 \\
\midrule
\textbf{BooF (ours)} & \textbf{0.959 $\pm$ 0.026} & \textbf{0.962 $\pm$ 0.026} & 0.853 $\pm$ 0.040 & \textbf{0.885 $\pm$ 0.035} \\
\midrule
\multicolumn{5}{c}{\textbf{BUS-BRA}} \\
\midrule
BD-StableNet \cite{bdstable} & 0.846 & 0.924 & 0.652 & 0.732 \\
SgmaFuse \cite{sgmafuse}     & 0.942 $\pm$ 0.015 & -- & -- & \textbf{0.946 $\pm$ 0.015} \\
BUS-BRA \cite{gomez2024bus}      & 0.931 $\pm$ 0.025 & 0.868 $\pm$ 0.014 & 0.859 $\pm$ 0.075 & -- \\
\midrule
\textbf{BooF (ours)} & \textbf{0.976 $\pm$ 0.004} & \textbf{0.949 $\pm$ 0.006} & \textbf{0.903 $\pm$ 0.019} & 0.899 $\pm$ 0.011 \\
\bottomrule[1pt]
\end{tabular}
\label{tab:main_results}
\vspace{-8pt}
\end{table}
\vspace{-5pt}
\subsubsection{Attention-Gated Multimodal Fusion Mechanism}
\vspace{-5pt}
Although the boot stage significantly improves the quality of MLLM-generated diagnostic descriptions, a second calibration step is required to mitigate residual hallucinations.
Inspired by the clinical practice of radiologists re-verifying imaging findings, we design an Attention-Gated Cross-Modality Fusion Module  (AGCFM) that selectively interacts with the most relevant BI-RADS descriptors, enabling the expert model to leverage semantic priors while suppressing hallucination noise \cite{wang2024improving}.
Formally, given the text embeddings \( \mathbf{T}_i \), the transformed image query and textual keys are defined as:
\begin{equation}
Q = W_Q \mathbf{I}, \quad
K = W_K [\mathbf{T}_1, \mathbf{T}_2, \dots, \mathbf{T}_C],
\end{equation}
where \( W_Q, W_K \) are learnable transformation matrices. The cross-modality attention matrix  \( A \in \mathbb{R}^{1 \times C} \) is computed as:

\begin{equation}
A = \text{softmax} \left( \frac{Q K^\top}{\sqrt{d}} \right).
\end{equation}
This quantifies the relevance of each textual attribute to the lesion’s visual features. The fused representation is then obtained by:

\begin{equation}
\mathbf{Z} = \mathbf{I} + A W_V [\mathbf{T}_1, \mathbf{T}_2, \dots, \mathbf{T}_C],
\end{equation}

where \( W_V \) projects text embeddings into the image feature space. The fused representation \( \mathbf{Z} \) is then passed through a fully connected layer to generate classification logits \( \hat{y} \), predicting lesion malignancy. The model is optimized using cross-entropy loss.
\label{sec:method}

\section{Experiments}
\label{sec:experiments}
\subsection{Experimental Setup} 
We evaluate our method on two publicly available breast ultrasound (BUS) datasets: {\bf BUS-BRA}~\cite{gomez2024bus} and {\bf BUSI}~\cite{BUSI}. The BUS-BRA dataset comprises 1268 and 607 benign and malignant BUS images, respectively, annotated with BI-RADS descriptors and biopsy-proven labels, while BUSI contains 437 benign BUS images and 210 malignant BUS images.  In implementation, we apply Gemini 2.5 Pro \cite{comanici2025gemini} as our base MLLM to generate diagnostic descriptions. The vision-expert model shares the same vision encoder and training pipeline as the Vision–Language Expert. Data preprocessing is standardized across all datasets to ensure consistency. Model performance is evaluated using six key metrics: AUC, Accuracy, Specificity, Precision, Recall, and F1-score. To ensure fair evaluation, all models are trained and tested under 5-fold cross-validation implemented in PyTorch and executed on NVIDIA A100 GPUs.
\vspace{-5pt}
\subsection{Ablation Studies on Boot Stage and Feedback Stage}
To evaluate the contribution of each component, we perform ablation studies on BUS-BRA using RadBERT \cite{radbert} as the text encoder and MedViT \cite{medvit} and ResNet-50 \cite{resnet} as visual encoders. Baselines — (i) Vision expert refers to the standalone vision classifier. (ii) Vision–Language Expert incorporates MLLM-generated descriptions into the expert model without any boot or feedback strategy. As shown in Table \ref{tab:ablation}, naively coupling an MLLM proves unreliable: for instance, on ResNet-50\cite{resnet}, AUC even drops below that of the vision-only baseline, suggesting that spurious or clinically inconsistent text can misguide the expert model. \textbf{Boot stage} — Introducing BI-RADS Lexicon Alignment constrains the MLLM to generate standardized, terminology-grounded descriptions rather than direct diagnostic judgments, leading to consistent performance gains. Further incorporating Vision Expert Prediction Priors Guidance yields the most substantial improvements—particularly in Recall and AUC—by grounding the MLLM’s reasoning in the vision expert’s benign–malignant predictions and effectively suppressing hallucinations. \textbf{Feedback stage} — The attention-gated multimodal fusion then integrates these enhanced textual descriptions with visual features while adaptively filtering semantic noise, delivering an additional and stable boost in performance. Across both backbones, the complete BooF configuration achieves the strongest results (bold in Table~\ref{tab:ablation}), with absolute improvements over the vision-only baseline of approximately 4.8\%–6.1\% in AUC, 5.7\%–7.4\% in Accuracy, and 14\%–16\% in Recall. These consistent gains confirm that BooF fosters a synergistic interaction between general MLLM reasoning and domain-specific expertise, thereby enhancing diagnostic reliability and ensuring stronger clinical alignment. Qualitatively, the MLLM \textit{self-corrects} erroneous vision priors, while feedback filters residual inconsistencies via attention fusion.
\vspace{-5pt}
\subsection{Comparisons with State-of-the-art Methods} 
We compare BooF with state-of-the-art methods on the BUSI \cite{BUSI} and BUS-BRA \cite{gomez2024bus} datasets (Table~\ref{tab:main_results}), which vary in scale and complexity. BooF outperforms all competitors, achieving the highest scores in most evaluation metrics. On BUSI, BooF achieves the best overall performance with an AUC of 0.959, Specificity of 0.962, and F1-score of 0.885, surpassing KRC-APM \cite{krc}, HoverTrans \cite{hovertrans}, and REAF \cite{reaf}. On BUS-BRA, BooF reaches an AUC of 0.976, Recall of 0.903, and F1-score of 0.899, outperforming BD-StableNet \cite{bdstable}, SgmaFuse \cite{sgmafuse}, and the baseline model. Beyond accuracy, BooF offers BI-RADS-aligned descriptions, providing interpretable, clinically grounded rationales that facilitate radiologist verification.
\vspace{-5pt}
\section{Conclusion}
\vspace{-5pt}
In this paper, we propose the Boot-and-Feedback framework, which enables a dynamic and bidirectional interaction between general Multimodal Large Language Models (MLLMs) and domain-specific expert models. By incorporating BI-RADS Lexicon Alignment and Vision Expert Prediction Priors Guidance in the boot stage, followed by multimodal attention fusion in the feedback stage, BooF significantly improves both diagnostic accuracy and interpretability in breast ultrasound analysis. Extensive experiments on two public datasets demonstrate that BooF surpasses state-of-the-art methods, offering superior diagnostic precision as well as clinically relevant decision rationales. The proposed framework is highly adaptable, making it applicable to other specialized classification tasks in medical diagnostics. Future research could extend the boot-and-feedback approach to multi-round iterative interactions, which may further enhance diagnostic performance and robustness.
\section{ACKNOWLEDGMENTS}
\label{sec:ack}
\vspace{-5pt}
This research was supported by the Australian Government through the Australian Research Council's DECRA funding scheme (Grant No. DE250100030) and Discovery Project funding scheme (Grant No. DP260100218).
\vfill\pagebreak

\label{sec:refs}

\small
\bibliographystyle{IEEEtran}
\bibliography{strings,refs}

@misc{WHO,
  author = {{World Health Organization}},
  title = {World Health Organization (WHO)},
  howpublished = {\url{https://www.who.int/}},
  year = {2024},
  note = {Accessed: 2024-05-12}
}

@inproceedings{resnet,
  title={Deep residual learning for image recognition},
  author={He, Kaiming and Zhang, Xiangyu and Ren, Shaoqing and Sun, Jian},
  booktitle={Proceedings of the IEEE conference on computer vision and pattern recognition},
  pages={770--778},
  year={2016}
}

@article{vit,
  title={An image is worth 16x16 words: Transformers for image recognition at scale},
  author={Dosovitskiy, Alexey and Beyer, Lucas and Kolesnikov, Alexander and Weissenborn, Dirk and Zhai, Xiaohua and Unterthiner, Thomas and Dehghani, Mostafa and Minderer, Matthias and Heigold, Georg and Gelly, Sylvain and others},
  journal={arXiv preprint arXiv:2010.11929},
  year={2020}
}

@book{BIRADS,
  author    = {{American College of Radiology}},
  title     = {BI-RADS: Breast Imaging Reporting and Data System},
  edition   = {5th},
  publisher = {American College of Radiology},
  address   = {Reston, VA},
  year      = {2013}
}

@article{kelly2010,
  title={Breast cancer detection: radiologists’ performance using mammography with and without automated whole-breast ultrasound},
  author={Kelly, Kevin M and Dean, Judy and Lee, Sung-Jae and Comulada, W Scott},
  journal={European radiology},
  volume={20},
  pages={2557--2564},
  year={2010},
  publisher={Springer}
}

@article{hovertrans,
  title={Hover-trans: Anatomy-aware hover-transformer for roi-free breast cancer diagnosis in ultrasound images},
  author={Mo, Yuhao and Han, Chu and Liu, Yu and Liu, Min and Shi, Zhenwei and Lin, Jiatai and Zhao, Bingchao and Huang, Chunwang and Qiu, Bingjiang and Cui, Yanfen and others},
  journal={IEEE Transactions on Medical Imaging},
  volume={42},
  number={6},
  pages={1696--1706},
  year={2023},
  publisher={IEEE}
}

@article{krc,
  title={KRC-APM: Key region cutting and artificial prior model for breast cancer recognition in ultrasound images},
  author={Lin, Yi and Wang, Haosen and Jiang, Jingchi},
  journal={Expert Systems with Applications},
  volume={257},
  pages={125092},
  year={2024},
  publisher={Elsevier}
}

@inproceedings{reaf,
  title={Reaf: Roi extraction and adaptive fusion for breast cancer diagnosis in ultrasound images},
  author={Zhang, Ziyang and Lim, Jia Wei and Zheng, Yuanxun and Chen, Bozhao and Chen, Dongxin and Lin, Yi},
  booktitle={2023 IEEE International Conference on Bioinformatics and Biomedicine (BIBM)},
  pages={3422--3429},
  year={2023},
  organization={IEEE}
}

@article{bdstable,
  title={BD-StableNet: a deep stable learning model with an automatic lesion area detection function for predicting malignancy in BI-RADS category 3--4A lesions},
  author={Qu, Hui and Chen, Guanglei and Li, Tong and Zou, Mingchen and Liu, Jiaxi and Dong, Canwei and Tian, Ye and Liu, Caigang and Cui, Xiaoyu},
  journal={Physics in Medicine \& Biology},
  volume={69},
  number={24},
  pages={245002},
  year={2024},
  publisher={IOP Publishing}
}

@inproceedings{fang2024aligning,
  title={Aligning Medical Images with General Knowledge from Large Language Models},
  author={Fang, Xiao and Lin, Yi and Zhang, Dong and Cheng, Kwang-Ting and Chen, Hao},
  booktitle={International Conference on Medical Image Computing and Computer-Assisted Intervention},
  pages={57--67},
  year={2024},
  organization={Springer}
}

@inproceedings{gao2024aligning,
  title={Aligning human knowledge with visual concepts towards explainable medical image classification},
  author={Gao, Yunhe and Gu, Difei and Zhou, Mu and Metaxas, Dimitris},
  booktitle={International Conference on Medical Image Computing and Computer-Assisted Intervention},
  pages={46--56},
  year={2024},
  organization={Springer}
}

@article{medvit,
  title={MedViT: a robust vision transformer for generalized medical image classification},
  author={Manzari, Omid Nejati and Ahmadabadi, Hamid and Kashiani, Hossein and Shokouhi, Shahriar B and Ayatollahi, Ahmad},
  journal={Computers in biology and medicine},
  volume={157},
  pages={106791},
  year={2023},
  publisher={Elsevier}
}

@article{comanici2025gemini,
  title={Gemini 2.5: Pushing the frontier with advanced reasoning, multimodality, long context, and next generation agentic capabilities},
  author={Comanici, Gheorghe and Bieber, Eric and Schaekermann, Mike and Pasupat, Ice and Sachdeva, Noveen and Dhillon, Inderjit and Blistein, Marcel and Ram, Ori and Zhang, Dan and Rosen, Evan and others},
  journal={arXiv preprint arXiv:2507.06261},
  year={2025}
}

@article{radbert,
  title={RadBERT: adapting transformer-based language models to radiology},
  author={Yan, An and McAuley, Julian and Lu, Xing and Du, Jiang and Chang, Eric Y and Gentili, Amilcare and Hsu, Chun-Nan},
  journal={Radiology: Artificial Intelligence},
  volume={4},
  number={4},
  pages={e210258},
  year={2022},
  publisher={Radiological Society of North America}
}

@inproceedings{yang2023generative,
  title={Generative speech recognition error correction with large language models and task-activating prompting},
  author={Yang, Chao-Han Huck and Gu, Yile and Liu, Yi-Chieh and Ghosh, Shalini and Bulyko, Ivan and Stolcke, Andreas},
  booktitle={2023 IEEE Automatic Speech Recognition and Understanding Workshop (ASRU)},
  pages={1--8},
  year={2023},
  organization={IEEE}
}

@article{brown2020language,
  title={Language models are few-shot learners},
  author={Brown, Tom and Mann, Benjamin and Ryder, Nick and Subbiah, Melanie and Kaplan, Jared D and Dhariwal, Prafulla and Neelakantan, Arvind and Shyam, Pranav and Sastry, Girish and Askell, Amanda and others},
  journal={Advances in neural information processing systems},
  volume={33},
  pages={1877--1901},
  year={2020}
}

@inproceedings{MedAlign2023,
  title={Medalign: A clinician-generated dataset for instruction following with electronic medical records},
  author={Fleming, Scott L and Lozano, Alejandro and Haberkorn, William J and Jindal, Jenelle A and Reis, Eduardo and Thapa, Rahul and Blankemeier, Louis and Genkins, Julian Z and Steinberg, Ethan and Nayak, Ashwin and others},
  booktitle={Proceedings of the AAAI Conference on Artificial Intelligence},
  volume={38},
  number={20},
  pages={22021--22030},
  year={2024}
}

@article{zhang2023biomedclip,
  title={Biomedclip: a multimodal biomedical foundation model pretrained from fifteen million scientific image-text pairs},
  author={Zhang, Sheng and Xu, Yanbo and Usuyama, Naoto and Xu, Hanwen and Bagga, Jaspreet and Tinn, Robert and Preston, Sam and Rao, Rajesh and Wei, Mu and Valluri, Naveen and others},
  journal={arXiv preprint arXiv:2303.00915},
  year={2023}
}

@article{BUSI,
  title={Dataset of breast ultrasound images},
  author={Al-Dhabyani, Walid and Gomaa, Mohammed and Khaled, Hussien and Fahmy, Aly},
  journal={Data in brief},
  volume={28},
  pages={104863},
  year={2020},
  publisher={Elsevier}
}

@article{gomez2024bus,
  title={BUS-BRA: a breast ultrasound dataset for assessing computer-aided diagnosis systems},
  author={G{\'o}mez-Flores, Wilfrido and Gregorio-Calas, Maria Julia and Coelho de Albuquerque Pereira, Wagner},
  journal={Medical physics},
  volume={51},
  number={4},
  pages={3110--3123},
  year={2024},
  publisher={Wiley Online Library}
}

@article{sgmafuse,
  title={Interpretable breast cancer diagnosis using histopathology and lesion mask as domain concepts conditional simulation ultrasonography},
  author={Dai, Guowei and Wang, Chaoyu and Tang, Qingfeng and Zhang, Yi and Dai, Duwei and Qiao, Lang and Yan, Jiaojun and Chen, Hu},
  journal={Information Fusion},
  pages={103343},
  year={2025},
  publisher={Elsevier}
}

@article{dai2026cinedub,
  title={CineDub: Scaling End-to-End Video Dubbing to Multi-Speaker Dialogues with Coherent Sound Effects},
  author={Dai, Yusheng and Wang, Kangdi and Gao, Baolong and Jiang, Yuxuan and Wang, Weiqiang and Ke, Qiuhong and Cai, Jianfei},
  journal={arXiv preprint arXiv:2608.15734},
  year={2026}
}

@inproceedings{wang2024improving,
  title={Improving multi-modal emotion recognition using entropy-based fusion and pruning-based network architecture optimization},
  author={Wang, Haotian and Du, Jun and Dai, Yusheng and Lee, Chin-Hui and Ren, Yuling and Liu, Yu},
  booktitle={ICASSP 2024-2024 IEEE International Conference on Acoustics, Speech and Signal Processing (ICASSP)},
  pages={11766--11770},
  year={2024},
  organization={IEEE}
}

@inproceedings{dai2023improving,
  title={Improving audio-visual speech recognition by lip-subword correlation based visual pre-training and cross-modal fusion encoder},
  author={Dai, Yusheng and Chen, Hang and Du, Jun and Ding, Xiaofei and Ding, Ning and Jiang, Feijun and Lee, Chin-Hui},
  booktitle={2023 IEEE International Conference on Multimedia and Expo (ICME)},
  pages={2627--2632},
  year={2023},
  organization={IEEE}
}
\end{document}